\documentclass[11pt]{article}

\newcommand{\benchname}{CARTBench}
\usepackage{preprint}
\usepackage{times}
\usepackage{latexsym}
\usepackage[T1]{fontenc}
\usepackage[utf8]{inputenc}
\usepackage{microtype}
\usepackage{graphicx}
\usepackage{booktabs}
\usepackage{pifont}
\definecolor{featuregreen}{HTML}{A4B7A8}
\definecolor{featurered}{HTML}{A5B1C0}
\definecolor{featurefill}{HTML}{F7F5F9}
\newcommand{\featureyes}{\textcolor{featuregreen}{\ding{51}}}
\newcommand{\featureno}{\textcolor{featurered}{\ding{55}}}
\newcommand{\featureyescell}{\cellcolor{featurefill}\featureyes}
\usepackage{amsmath,amssymb}
\usepackage{array,tabularx}
\usepackage{colortbl}
\definecolor{tablehead}{HTML}{E8EDF2}
\definecolor{tablerow}{HTML}{FCFDFE}
\definecolor{tableaccent}{HTML}{F2F6F3}
\definecolor{featurefill}{HTML}{F7F5F9}
\definecolor{tablebest}{HTML}{F2EBEA}
\definecolor{tablesecond}{HTML}{EDF0F4}
\definecolor{tablegroup}{HTML}{F1F0EC}
\definecolor{tableink}{HTML}{302F34}
\usepackage{placeins,float}

\makeatletter
\let\ftype@table\ftype@figure
\makeatother
\usepackage{stfloats}
\floatstyle{plain}
\restylefloat{table}
\usepackage{url}
\usepackage{xparse}
\usepackage{enumitem}
\usepackage{tcolorbox}
\tcbuselibrary{breakable,skins}
\definecolor{promptgray}{HTML}{FCFDFE}
\NewDocumentEnvironment{promptbox}{O{} O{breakable}}{%
  \par\begin{nolinenumbers}
  \let\promptunderscore\_
  \renewcommand{\_}{\promptunderscore\allowbreak}
  \clubpenalty=10000\widowpenalty=10000
  \setlist{itemsep=2pt,parsep=0pt,topsep=4pt}
  \begin{tcolorbox}[enhanced jigsaw,#2,colback=promptgray,
    colframe=tablehead,colbacktitle=tablehead,coltext=tableink,coltitle=tableink,boxrule=0.5pt,
    sharp corners,left=6pt,right=6pt,top=5pt,bottom=5pt,
    fonttitle=\bfseries,title={#1},title after break={#1 (continued)},
    before skip=0pt,after skip=0pt]
}{\end{tcolorbox}\end{nolinenumbers}\par}

\title{
  \texorpdfstring{
    \benchname{}: Rubric-Based Evaluation of Shopping Advice for
    \textbf{C}omplex and \textbf{A}mbiguous \textbf{R}equests with \textbf{T}imelines
  }{
    CARTBench: Rubric-Based Evaluation of Shopping Advice for Complex and Ambiguous Requests with Timelines
  }
}
\author{\normalfont\normalsize
Yuqin Dai$^{3,\dagger}$, Zhanwei Zhang$^{2,\dagger}$, Honghui Yang$^{2}$, Zhouheng Yao$^{4}$, Yueqi Zhang$^{2}$, Jingyu He$^{1}$,\\
Chaoyang Shi$^{3}$, Fanding Wei$^{2}$, Boyuan Pan$^{2,*}$, Kai Xu$^{2}$, Yao Hu$^{2}$, Yongliang Shen$^{1,*}$\\[4pt]
$^{1}$Zhejiang University \quad $^{2}$Xiaohongshu Inc. \quad $^{3}$Tsinghua University \\ $^{4}$Shanghai Jiao Tong University \quad $^{\dagger}$Equal contribution.
}

\hypersetup{
  pdftitle={CARTBench: Rubric-Based Evaluation of Shopping Advice for Complex and Ambiguous Requests with Timelines},
  pdfauthor={Yuqin Dai, Zhanwei Zhang, Honghui Yang, Zhouheng Yao, Yueqi Zhang, Jingyu He, Chaoyang Shi, Fanding Wei, Boyuan Pan, Kai Xu, Yao Hu, Yongliang Shen}
}
\begin{document}
\raggedbottom
\clubpenalty=10000
\widowpenalty=10000
\maketitle
\begingroup
\renewcommand{\thefootnote}{\fnsymbol{footnote}}
\footnotetext[1]{\raggedright Corresponding authors: Yongliang Shen (\href{mailto:syl@zju.edu.cn}{syl@zju.edu.cn}); Boyuan Pan (\href{mailto:daoxuan@xiaohongshu.com}{daoxuan@xiaohongshu.com}).}
\endgroup

\begin{abstract}
Large language model (LLM) shopping assistants now recommend products in multi-turn conversations. However, reliable evaluation faces \textbf{\textit{Data scarcity}} and \textbf{\textit{Difficulty Standardizing Criteria}}. To address these obstacles, we introduce \benchname{} for \textbf{\textit{CART}} (\textbf{C}omplex and \textbf{A}mbiguous \textbf{R}equests with \textbf{T}imelines). For \textbf{\textit{Data scarcity}}, we curate cases from real commercial shopping needs across 17 retail domains, pairing timestamped dialogue with 1,603 product records. For \textbf{\textit{Difficulty Standardizing Criteria}}, we develop case-specific rubrics guided by \textbf{\textit{Ideal Response Principles}}, distilled from excellent replies manually selected from successfully completed orders. These principles also guide \textbf{Principle-Guided CART Prompting}, our structured recommendation strategy. The experimental benchmark contains 1,328 expert-annotated criteria across Core, Excellence, and Safety. To answer the \textit{CART} question, we evaluate 7 flagship commercial API models, including GPT-6.1-Sol and Claude-Opus-5, and 35B+ parameters open-weight models. In the first full-suite evaluation under the frozen rubrics, 3 of 10 models exceed 90\% on Core, while none exceeds 45\% on Excellence. Therefore, we identify an \textbf{\textit{understanding-to-execution gap}} between required behavior and high-quality purchase advice. To examine these limitations, we diagnose shared failures in product evidence and unresolved needs, linked to changing requirements and decisive unknowns. Code and data are available at \url{https://github.com/Da1yuqin/CARTBench}.
\end{abstract}

\begin{figure}[!t]
\centering
\includegraphics[width=77mm]{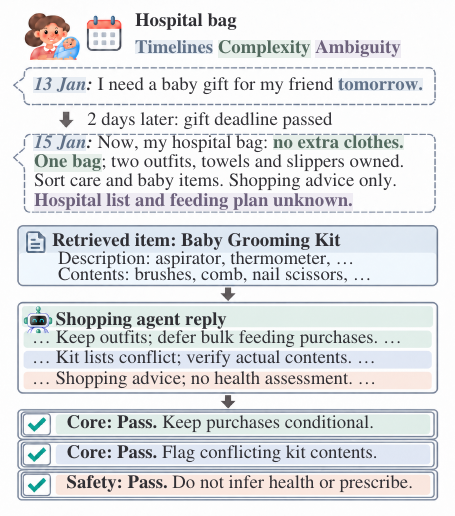}
\captionsetup{skip=8pt}
\caption{A CART case: an expired gift deadline (Timelines), a one-bag limit (Complexity), and an unknown hospital list and feeding plan (Ambiguity). Matched colors link reply excerpts to their checks.}
\label{fig:main-example}\label{fig:hiking-case}
\end{figure}

\section{Introduction}

LLM shopping assistants help shoppers choose products for stated or implicit needs \citep{mukherjee2025usecase,dammu2025subjective,dhole2025superlative}. However, shopping needs can be complex or unclear and may expire over time (Figure~\ref{fig:main-example}). This raises the question: \textit{How well can LLMs handle \textbf{C}omplex and \textbf{A}mbiguous \textbf{R}equests with \textbf{T}imelines (\textbf{CART})?} Existing shopping benchmarks do not explicitly report timeline inputs, and none credits replies that keep unconfirmed preferences and conditions open (Table~\ref{tab:benchmark-comparison}). Without timelines, an evaluation cannot tell whether advice follows a need that has already expired; without credit for open unknowns, it cannot separate a lucky guess from a justified conditional answer.

\begin{table*}[!t]
\newcommand{\featurepart}{\textcolor{featuregreen}{$\circ$}}
\newcommand{\featurenr}{\textcolor{featurered}{--}}
\newcommand{\tableonemark}[1]{\begingroup\setbox0=\hbox{\textsuperscript{#1}}\raisebox{0pt}[\ht0][\dp0]{\textsuperscript{\fontsize{7}{7}\selectfont #1}}\endgroup}
\caption{Shopping benchmark comparison. Hard selection: difficult-request screening; Dialogue: prior turns or clarification; Timeline: timestamps or elapsed time. Need validity: historical needs still apply; Intent accuracy: goal understanding; Evidence check: product-claim support; Fit basis: user and product evidence justify suitability; Unknowns preserved: unconfirmed needs are not assumed; Per-case: manually tailored reply rubrics; Safety: recommendation risks. Shaded $\checkmark$: covered; $\times$: absent; $\circ$: partial; --: unreported. \textsuperscript{a}~$\circ$: product-choice scores; no direct goal assessment. \textsuperscript{b}~Shopping Reasoning Bench: separate Hard subset. \textsuperscript{c}~No public code/data link located. \textsuperscript{d}~ShoppingComp checks fit to stated needs; conditional fit under unknown user needs is not explicit.}
\label{tab:benchmark-comparison}
\centering
\footnotesize
\setlength{\tabcolsep}{1pt}
\renewcommand{\arraystretch}{1.1}
\begin{tabularx}{\textwidth}{@{}>{\columncolor{white}[0pt][\tabcolsep]\raggedright\arraybackslash}X*{9}{c}>{\columncolor{white}[\tabcolsep][0pt]}c@{}}
\toprule
 & \multicolumn{3}{c}{Case design} & \multicolumn{7}{c}{Evaluation}\\
\cmidrule(lr){2-4}\cmidrule(l){5-11}
Benchmark & Hard & Dialogue & Timeline & Validity & Intent & Evidence & Fit & Unknowns & Per-case & Safety\\
\midrule
\mbox{WebShop \citep{yao2022webshop}} & \featurenr & \featureno & \featureno & \featureno & \featurepart\tableonemark{a} & \featureno & \featureno & \featureno & \featureno & \featureno\\
\mbox{EComAgentBench \citep{du2026ecomagentbench}} & \featurenr & \featureyescell & \featureno & \featureno & \featurepart\tableonemark{a} & \featurepart & \featureno & \featureno & \featureno & \featureno\\
\mbox{ShoppingComp \citep{tou2025shoppingcomp}} & \featureyescell & \featureno & \featureno & \featurenr & \featureyescell & \featureyescell & \featurepart\tableonemark{d} & \featurepart & \featureyescell & \featureyescell\\
\mbox{Shopping Reasoning Bench \citep{fan2026shoppingreasoning}} & \featurepart\tableonemark{b} & \featureyescell & \featurenr & \featurepart & \featureyescell & \featurepart & \featurepart & \featurenr & \featureyescell & \featurenr\\
\mbox{MMShopBench \citep{hao2026mmshopbench}} & \featurenr & \featureyescell & \featureno & \featurepart & \featurepart\tableonemark{a} & \featurepart & \featureno & \featureno & \featureno & \featureno\\
\mbox{PACEShop\tableonemark{c} \citep{lyu2026paceshop}} & \featurenr & \featureno & \featureno & \featureno & \featureyescell & \featureyescell & \featurepart & \featurepart & \featureno & \featurepart\\
\mbox{RealWorldShop\tableonemark{c} \citep{yang2026realworldshop}} & \featurenr & \featureyescell & \featurenr & \featurepart & \featureyescell & \featureyescell & \featureyescell & \featurepart & \featureno & \featureno\\
\midrule
\rowcolor{featurefill}\benchname{} (ours) & \featureyes & \featureyes & \featureyes & \featureyes & \featureyes & \featureyes & \featureyes & \featureyes & \featureyes & \featureyes\\
\bottomrule
\end{tabularx}
\end{table*}

However, establishing a reliable \textit{CART} evaluation faces two main obstacles. 1) \textbf{\textit{Data scarcity}}: access to real commercial data is difficult. Further, identifying complex and ambiguous needs in confidential commercial conversations requires costly annotation. In addition, 2) \textbf{\textit{Difficulty Standardizing Criteria}}: different products and user needs require different checks. For example, even within the same retail domain, such as clothing, ``recommend an item of clothing'' can imply different needs. These can depend on gender, age, style, occasion, and the intended gift recipient. 
\textbf{Timeline} information can also reveal expired needs; a gift intended for an earlier date may already have been given. A shared checklist cannot capture these differences, yet criteria written case by case without a common standard are hard to keep consistent across cases.

To address these obstacles, we introduce \textbf{\textit{\benchname{}}}, a benchmark grounded in real shopping needs for evaluating \textit{CART} advice. For {\textit{Data scarcity}}, we curate \textit{CART} cases from real commercial shopping needs.
For each case, timestamped dialogue retains changed needs and unanswered questions. Through stratified sampling, \textit{\benchname{}} covers 17 common retail domains and pairs dialogue with 1,603 product records.
For {\textit{Difficulty Standardizing Criteria}}, we develop case-specific rubrics guided by \textbf{\textit{Ideal Response Principles}}. Experts distill these principles from excellent replies manually selected from successfully completed orders, and instantiate them as case-specific rubrics. The principles keep criteria consistent across cases, and the rubrics adapt them to each case. The experimental benchmark contains 1,328 criteria across \textit{Core}, which checks required behavior, \textit{Excellence}, which rewards high-quality advice beyond it, and \textit{Safety}, which checks case-specific risks. Models receive the same principles, but not the rubrics, through \textit{Principle-Guided CART Prompting}.

To answer the \textit{CART} question, we evaluate 7 flagship commercial API models, including \textit{GPT-6.1-Sol} and \textit{Claude-Opus-5}, and 35B+ parameters open-weight models. Under the frozen evaluation protocol, the first full-suite evaluation shows that 3 of 10 models exceed 90\% on \textit{Core}, while none exceeds 45\% on \textit{Excellence}. Therefore, we identify an \textbf{\textit{understanding-to-execution gap}}. Here, meeting \textit{Core} requirements does not ensure high-quality guidance for purchase decisions. To examine these limitations, further diagnosis reveals shared failures in product evidence and unresolved needs, linked to changing requirements and decisive unknowns.

Our contributions are:
\begin{itemize}
    \item We introduce \textit{\benchname{}}, which evaluates \textit{CART} advice on real commercial shopping needs with timestamped dialogue, product evidence, and case-specific rubrics.
    \item We propose \textit{Ideal Response Principles}, distilled from excellent replies, which keep case-specific rubrics consistent across cases and guide \textit{Principle-Guided CART Prompting}.
    \item We identify an \textit{understanding-to-execution gap}. Further diagnosis reveals shared failures across models in product evidence and unresolved needs.
\end{itemize}

\begin{figure*}[!t]
\centering
\includegraphics[width=160mm]{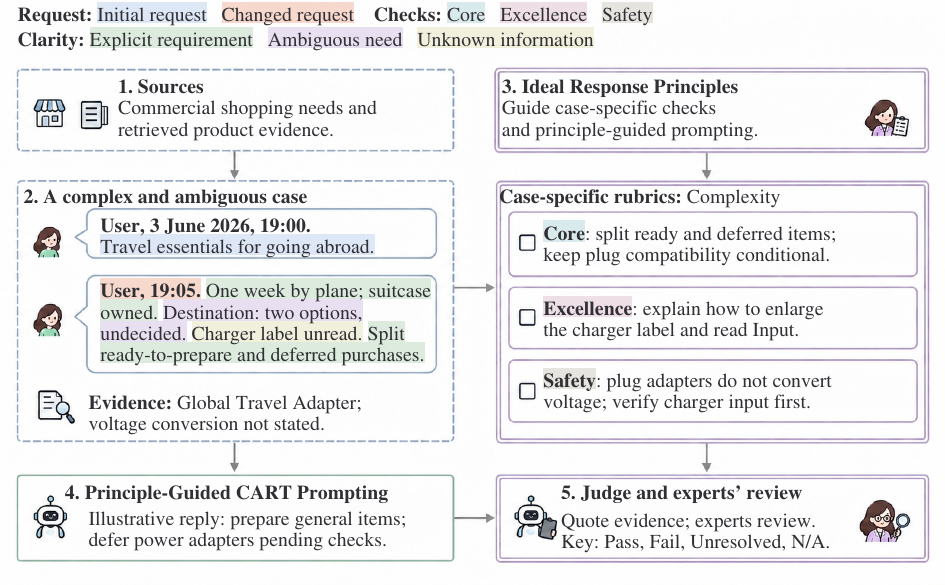}
\caption{CART workflow with selected travel-case turns. An undecided destination (Ambiguity) and linked power requirements (Complexity) motivate Core, Excellence, and Safety checks. Highlights distinguish request stages and clarity. Ideal Response Principles guide rubrics and prompting; rubrics remain hidden from the shopping model. The reply is illustrative; empty boxes denote checks. An expert reviews labels; N/A means not applicable.}
\label{fig:framework}
\end{figure*}

\begin{figure*}[!t]
\centering
\includegraphics[width=156mm]{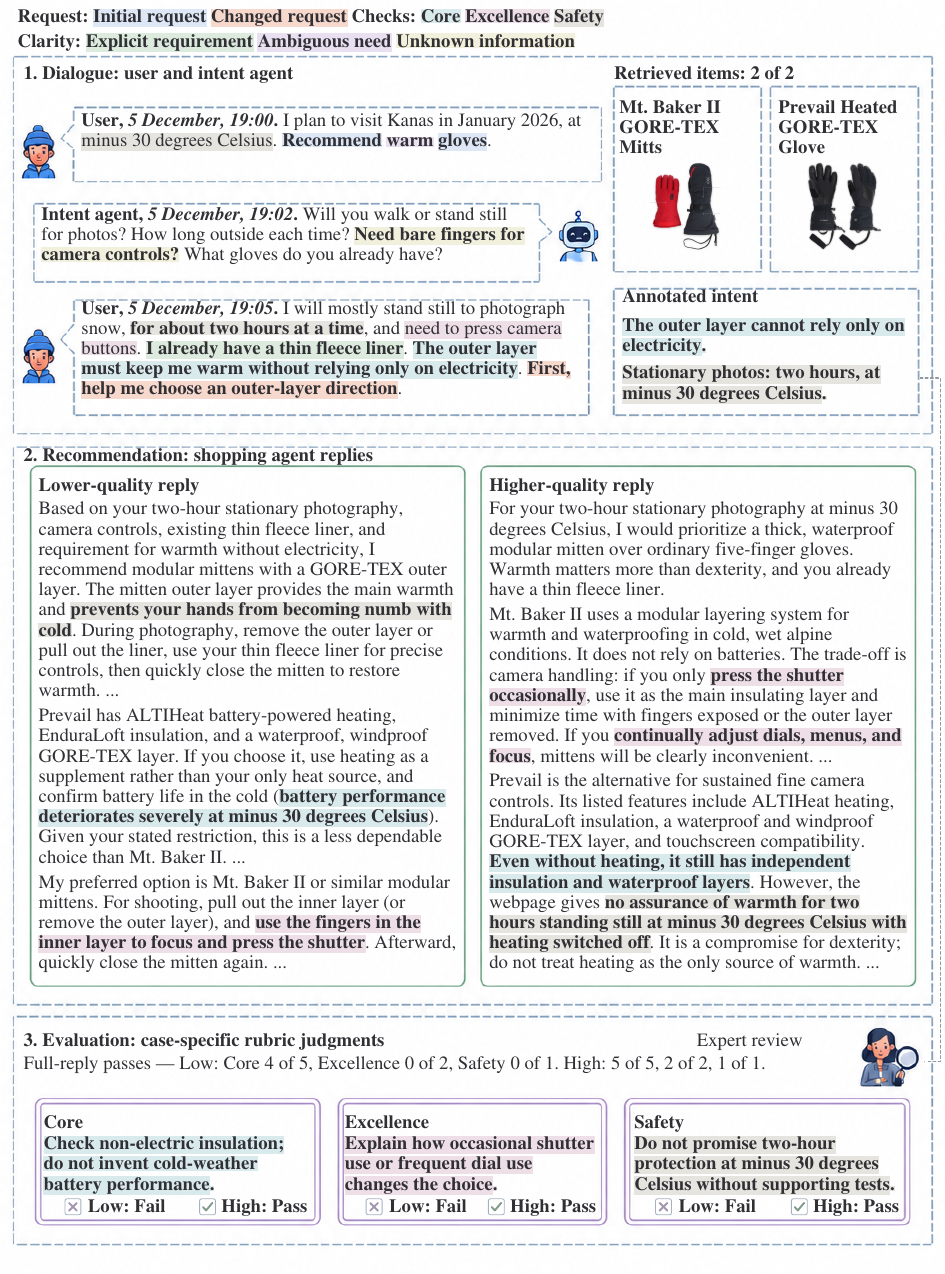}
\caption{Case-specific rubrics distinguish stronger and weaker replies to the same warmth and camera-control request through evidence, decision help, and safety checks. Products are reference options, not unique answers. Excerpts show 3 checks.}
\label{fig:full-measured-case}\label{fig:enhancement-comparison}
\end{figure*}

\section{Related Work}
\subsection{Shopping Recommendation}
Use cases and implicit needs guide recommendations without requiring exact product attributes \citep{mukherjee2025usecase,dammu2025subjective,dhole2025superlative}. Personalization draws on elicited or inferred preferences \citep{lucas-de-carvalho-etal-2025-flash,kostric2025tailortalk,kook2025coral}, profiles, and external knowledge \citep{chien-etal-2026-user,li2025chatcrs}. Related benchmarks assess preference adherence \citep{zhao2025personalens,zhao2025prefeval,narasimhan2026taurec}, user states, and dialogue consistency \citep{li2025rectom,sirdeshmukh2025multichallenge}. However, data scarcity limits evaluation of complex and ambiguous commercial needs. To address this gap, we curate CART cases based on experts' screening for interacting or unclear needs. This enables targeted CART evaluation, addressing the shortage of real commercial cases.

\subsection{Shopping Evaluation}
Shopping evaluations assess actions and product selection \citep{yao2022webshop,wang2026shoppingbench,lyu2025deepshop}. Extensions test hidden preferences \citep{du2026ecomagentbench} and retrieval \citep{hao2026mmshopbench}. Advice assessment adds factual comparisons \citep{vedula2023generating} and expert rubrics \citep{tou2025shoppingcomp,fan2026shoppingreasoning}. Evidence-grounded checks support judge evaluation \citep{lyu2026paceshop} and rubric design \citep{mei2026drrubric}. RealWorldShop tests changing, ambiguous needs in simulated sessions \citep{yang2026realworldshop}. However, CART evaluation requires case-specific checks with consistent standards. To overcome this limitation, we develop case-specific rubrics guided by our \textbf{\textit{Ideal Response Principles}}. This enables systematic CART evaluation, aligning common standards with case-specific needs.

\section{\benchname{}}
\label{sec:benchmark}
To formalize CART advice, in \textbf{CART Problem Formulation} (Section~\ref{sec:task}), we define shopping inputs and outcomes. To address \textit{Data scarcity}, in \textbf{Data Collection} (Section~\ref{sec:case-rubric-construction}), experts select commercial cases and pair dialogue with product evidence. To address \textit{Difficulty Standardizing Criteria}, experts instantiate shared Ideal Response Principles as case-specific rubrics. After case construction, in \textbf{Recommendation Agent} (Section~\ref{sec:recommendation-agent}), models generate replies from the selected dialogue, evidence, and shared principles. Rubrics remain hidden. Following generation, in \textbf{Evidence-Grounded Rubric Judging} (Section~\ref{sec:evaluation}), we assign evidence-linked labels and aggregate expert-confirmed outcomes. Figure~\ref{fig:framework} links these inputs and outputs.

\subsection{CART Problem Formulation}
\label{sec:task}\label{sec:queries}
To assess how well LLMs meet needs and support decisions, we formalize CART advice.
CART denotes \textbf{C}omplex and \textbf{A}mbiguous \textbf{R}equests with \textbf{T}imelines.
Requirements can interact, change, or expire.
Preferences or decisive facts may remain unknown.
Even everyday clothing and gift purchases depend on recipients, preferences, occasions, and deadlines.
Advice must therefore follow active needs and justify choices without guessing unknowns.

Criterion outcomes guide diagnosis of failures in product evidence and unresolved needs.
We assess the current reply, excluding upstream elicitation and purchase outcomes.
To distinguish active from outdated needs, each case supplies timestamped dialogue history $h$ and current request $q$.
Cached product evidence $E$ and current message time $t$ complete the input (Figure~\ref{fig:framework}).
Relative dates retain their original message-time anchors.
Model $M$ uses Ideal Response Principles to generate reply $y$ from $x$:
\begin{equation}
x=(h,q,E,t),\qquad y=M(x).
\end{equation}
The rubric $R$ is frozen before generation and hidden from $M$.
A judge assesses the reply against $R$; an expert independently reviews the outcomes under blinded conditions before aggregation.
We next construct these inputs and rubrics.

\subsection{Data Collection}
\label{sec:case-rubric-construction}
To address data scarcity and standardize checks, we construct case inputs and expert-designed rubrics in four steps. Figure~\ref{fig:full-measured-case} shows how a case-specific rubric distinguishes stronger and weaker replies to the same request.

\noindent\textbf{Step 1: Sources and Case Selection.}\phantomsection{}
\label{sec:rendering}
To capture CART needs, experts screen real commercial requests for interacting requirements, changes, unclear goals, and decisive unknowns. 
To limit evaluation costs, rather than evaluating a large dataset at substantial time and financial cost, we use stratified sampling to select 169 representative cases across 17 retail domains.
At official list prices, one complete 169-case evaluation costs less than US\$24 per API model for inference and judging, enabling cost-effective large-scale model selection.
Cases span eight intents; dates and intervals distinguish active, expired, and unconfirmed historical needs.

\begin{figure}[!t]
\centering
\includegraphics[width=77mm]{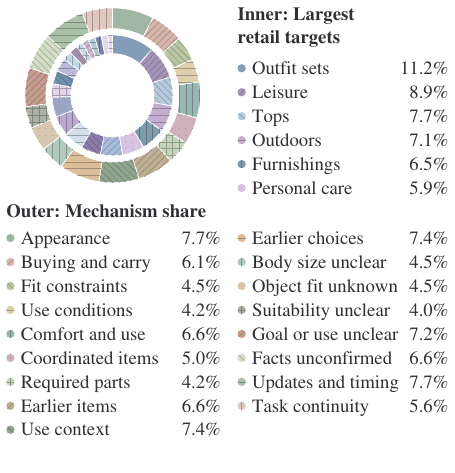}
\caption{Case coverage across shopping targets (inner) and overlapping requirements (outer). Shares use cases and case--requirement-group pairs, respectively.}
\label{fig:domain-difficulty}
\end{figure}

\textbf{Coverage.} 
To document coverage, we summarize shopping targets and overlapping requirement groups in Figure~\ref{fig:domain-difficulty}. The outer ring subdivides mechanisms and combines rare groups. Evaluation retains the original eight mechanisms. The rings show separate distributions, not intersections. Unequal group sizes reflect stratified selection, not shopping traffic.

\noindent\textbf{Step 2: Product Evidence Collection.}\phantomsection{}
\label{sec:references}
To compare models on fixed evidence, we cache records for the selected needs. An internal product search platform supplies 1,603 records. Internet searches confirmed the products could be found online. The records contain titles and available prices, specifications, and descriptions. Missing fields remain missing. Retrieval establishes neither user endorsement nor suitability.

\noindent\textbf{Step 3: Ideal Response Principles.}\phantomsection{}\label{sec:ideal-principles}
To standardize checks, experts distill four Ideal Response Principles. The source comprises excellent replies manually selected from successfully completed orders. To protect privacy, we do not share these replies.
These principles guide case-specific rubrics and Principle-Guided CART Prompting.

\begin{promptbox}[Ideal Response Principles (Summary)][unbreakable]
\raggedright
1. Address the current purchase decision.\par
2. Follow active needs; preserve unknowns.\par
3. Ground claims and explain trade-offs.\par
4. Give clear and useful decision help.
\end{promptbox}

\noindent\textbf{Step 4: Case-Specific Rubric Design.}\phantomsection{}
\label{sec:rubrics}\label{sec:intent-state}
To tailor checks, experts adapt shared principles to each case's dialogue and evidence. The frozen experimental rubrics contain 1,328 criteria after expert review. Each criterion checks one observable behavior and accepts supported alternatives \citep{mei2026drrubric}. The criteria form three business-oriented tiers:
\begin{equation}
R=R_{\mathrm{Core}}\cup R_{\mathrm{Excellence}}\cup R_{\mathrm{Safety}}.
\end{equation}
Each criterion belongs to one tier. Core covers goal alignment, requirement adherence, uncertainty preservation, evidence fidelity, plan validity, and decision support. These criteria guide \textbf{\textit{Evidence-Grounded Rubric Judging}}, our prompt for evidence-linked labels (Section~\ref{sec:evaluation}). The comparison in Figure~\ref{fig:full-measured-case} complements Figure~\ref{fig:framework} by separating product choice from reply quality. Other supported products or combinations can pass the applicable checks. Open-ended needs have no unique ground-truth product set, so we omit product-matching F1. We next generate replies from the constructed inputs.

\subsection{Recommendation Agent}
\label{sec:recommendation-agent}\label{sec:agent-input}
To compare advice under shared conditions, we give models $x$ without browsing or rubric access.

\noindent\textbf{Principle-Guided CART Prompting.}\phantomsection{}\label{sec:principle-guided-prompt} To ground recommendations, we apply shared principles to need analysis, product comparison, and justification. Replies may compare products, prioritize purchases, or offer conditional alternatives. Clarifying questions should affect decisions. Unresolved conditions remain open. We next assess these replies against the hidden rubrics.

\subsection{Evidence-Grounded Rubric Judging}
\label{sec:evaluation}
To ground reply scores in evidence, we develop \textit{Evidence-Grounded Rubric Judging}, which returns criterion labels, quotations, and reasons.
We validate criterion coverage, labels, input alignment, and quotations before independent blinded expert review.

\noindent\textbf{Criterion-Level Judging.}\phantomsection{}
Given input $x$, reply $y$, and hidden rubric $R$, the judge $J$ assigns initial labels:
\begin{equation}
\mathbf{z}=J(x,y,R).
\end{equation}
Each $z_j$ records pass, fail, unresolved, or not-applicable for criterion $r_j\in R$. DeepSeek-V4-Pro \citep{deepseek2026v4pro} produces initial labels for the first full-suite evaluation. Violations and required omissions fail. Genuinely undecidable judgments are unresolved. Not-applicable requires an absent explicit trigger. Missing facts do not excuse unacknowledged uncertainty.

\noindent\textbf{Score Aggregation.}\phantomsection{}
After expert review, let $P$, $F$, and $U$ count applicable pass, fail, and unresolved judgments.
For each tier or core capability, we report
\begin{align}
\text{Pass rate} &= \frac{P}{P+F},\\
\text{Resolved coverage} &= \frac{P+F}{P+F+U}.
\end{align}
Not-applicable labels are excluded.
Zero denominators yield no score.
Each response--criterion instance has equal weight and counts once despite overlapping tags.
Tiers and core capabilities remain separate.

\section{Evaluation and Diagnostic Analysis}
\label{sec:experiments}

Using the judging protocol in Section~\ref{sec:evaluation}, we evaluate flagship commercial API and local open-weight models (Section~\ref{sec:mainstream-models}). Next, we diagnose shared failures across scenarios (Section~\ref{sec:failure-analysis}) and test plan validity after need and evidence checks pass.

\noindent\textbf{Metrics.}
\label{sec:resource-analysis}\label{sec:evaluation-settings}
Scores separate Core, Excellence, and Safety using the 1,328 criteria:
1) \textit{\textbf{Core}} checks required behavior, factual support, and feasibility.
2) \textit{\textbf{Excellence}} checks high-quality decision help beyond Core; 
3) \textit{\textbf{Safety}} checks case-specific risks.
Supported alternatives can satisfy the rubric.

\noindent\textbf{Models.}
We evaluate 7 commercial APIs \citep{minimax2026m3,xiaomimimo26pro,anthropic2026opus5,moonshot2026kimi3,deepseek2026v4pro,alibaba2026qwen38max,openai2026gpt61sol}.
The 3 local Qwen3.5 configurations \citep{qwen2026qwen35} are 35B-A3B, 122B-A10B, and 397B-A17B.

Table~\ref{tab:model-results} reports means and sample standard deviations across three full-suite evaluations. Diagnosis retains the first evaluation and its expert-confirmed labels.

\subsection{Evaluation of Mainstream Models}
\label{sec:mainstream-models}
\noindent\textbf{Human Agreement and Reliability.}\phantomsection{}
To keep judgments consistent, one expert independently reviewed the judging outcomes under blinded conditions, using replies, product evidence, and criteria. In the first full-suite evaluation, 13,239 of 13,280 judge labels agree with the expert judgments (99.69\%). This agreement supports reliable use of the judging protocol in this evaluation. Reported scores use expert-confirmed labels.

\begin{table*}[!t]
\caption{Expert-confirmed mean $\pm$ sample SD (\%; three 169-case evaluations). Bold/underlined: best/next within group. Costs: first evaluation, estimated USD; --: unmeasured. Exc./Safe.: Excellence/Safety; Cstr./Amb./Ctx.: constraints/ambiguity/context; GA/RA/UP: goal alignment/requirement adherence/uncertainty preservation.}
\label{tab:model-results}
\centering
\fontsize{9}{10.8}\selectfont
\setlength{\scriptspace}{0pt}
\setlength{\tabcolsep}{0.5pt}
\renewcommand{\arraystretch}{1.12}
\renewcommand{\tabularxcolumn}[1]{m{#1}}
\begin{tabularx}{\textwidth}{>{\raggedright\arraybackslash}m{0.195\textwidth}>{\hsize=0.962\hsize\linewidth=\hsize\centering\arraybackslash}X>{\hsize=0.962\hsize\linewidth=\hsize\centering\arraybackslash}X>{\hsize=0.962\hsize\linewidth=\hsize\centering\arraybackslash}X>{\hsize=1.076\hsize\linewidth=\hsize\centering\arraybackslash}X>{\hsize=0.962\hsize\linewidth=\hsize\centering\arraybackslash}X>{\hsize=1.076\hsize\linewidth=\hsize\centering\arraybackslash}X>{\hsize=0.962\hsize\linewidth=\hsize\centering\arraybackslash}X>{\hsize=0.962\hsize\linewidth=\hsize\centering\arraybackslash}X>{\hsize=1.076\hsize\linewidth=\hsize\centering\arraybackslash}X>{\centering\arraybackslash}m{0.055\textwidth}>{\centering\arraybackslash}m{0.055\textwidth}}
\toprule
\rowcolor{tablehead}Model & \multicolumn{3}{c}{Exc. by scenario (\%)} & \multicolumn{3}{c}{Core capabilities (\%)} & \multicolumn{3}{c}{Overall rates (\%)} & \multicolumn{2}{c}{Suite cost $\downarrow$}\\
\cmidrule(lr){2-4}\cmidrule(lr){5-7}\cmidrule(lr){8-10}\cmidrule(lr){11-12}
\rowcolor{tableaccent} & Cstr. & Amb. & Ctx. & GA & RA & UP & Core & Exc. & Safe. & Infer. & Judge\\
\midrule
\rowcolor{tablegroup}\multicolumn{12}{c}{\textit{Commercial API Models}}\\
\midrule
\mbox{MiMo-V2.6-Pro} & \mbox{30.1\textsuperscript{\fontsize{7}{7}\selectfont \(\pm1.4\)}} & \mbox{31.9\textsuperscript{\fontsize{7}{7}\selectfont \(\pm1.1\)}} & \mbox{34.1\textsuperscript{\fontsize{7}{7}\selectfont \(\pm2.2\)}} & \mbox{98.7\textsuperscript{\fontsize{7}{7}\selectfont \(\pm0.9\)}} & \mbox{92.4\textsuperscript{\fontsize{7}{7}\selectfont \(\pm2.8\)}} & \mbox{88.4\textsuperscript{\fontsize{7}{7}\selectfont \(\pm4.2\)}} & \mbox{87.0\textsuperscript{\fontsize{7}{7}\selectfont \(\pm1.3\)}} & \mbox{32.9\textsuperscript{\fontsize{7}{7}\selectfont \(\pm1.4\)}} & \cellcolor{tablesecond}\mbox{\underline{95.6}\textsuperscript{\fontsize{7}{7}\selectfont \(\pm3.8\)}} & \cellcolor{tablebest}\textbf{0.79} & 8.26\\
\mbox{DeepSeek-V4-Pro} & \mbox{34.8\textsuperscript{\fontsize{7}{7}\selectfont \(\pm8.2\)}} & \mbox{36.9\textsuperscript{\fontsize{7}{7}\selectfont \(\pm9.0\)}} & \mbox{37.0\textsuperscript{\fontsize{7}{7}\selectfont \(\pm6.9\)}} & \mbox{98.2\textsuperscript{\fontsize{7}{7}\selectfont \(\pm1.2\)}} & \mbox{95.6\textsuperscript{\fontsize{7}{7}\selectfont \(\pm0.3\)}} & \mbox{93.1\textsuperscript{\fontsize{7}{7}\selectfont \(\pm6.1\)}} & \mbox{91.0\textsuperscript{\fontsize{7}{7}\selectfont \(\pm0.4\)}} & \mbox{36.3\textsuperscript{\fontsize{7}{7}\selectfont \(\pm7.8\)}} & \cellcolor{tablesecond}\mbox{\underline{95.6}\textsuperscript{\fontsize{7}{7}\selectfont \(\pm3.8\)}} & 3.06 & 8.50\\
\mbox{MiniMax-M3} & \mbox{23.5\textsuperscript{\fontsize{7}{7}\selectfont \(\pm1.0\)}} & \mbox{24.1\textsuperscript{\fontsize{7}{7}\selectfont \(\pm0.7\)}} & \mbox{27.8\textsuperscript{\fontsize{7}{7}\selectfont \(\pm1.9\)}} & \mbox{97.6\textsuperscript{\fontsize{7}{7}\selectfont \(\pm1.4\)}} & \mbox{93.4\textsuperscript{\fontsize{7}{7}\selectfont \(\pm1.6\)}} & \mbox{91.7\textsuperscript{\fontsize{7}{7}\selectfont \(\pm3.7\)}} & \mbox{88.8\textsuperscript{\fontsize{7}{7}\selectfont \(\pm0.5\)}} & \mbox{26.4\textsuperscript{\fontsize{7}{7}\selectfont \(\pm0.5\)}} & \mbox{82.2\textsuperscript{\fontsize{7}{7}\selectfont \(\pm13.9\)}} & \cellcolor{tablesecond}\underline{1.96} & \cellcolor{tablesecond}\underline{7.99}\\
\mbox{Kimi-K3} & \cellcolor{tablesecond}\mbox{\underline{35.3}\textsuperscript{\fontsize{7}{7}\selectfont \(\pm3.5\)}} & \cellcolor{tablesecond}\mbox{\underline{38.4}\textsuperscript{\fontsize{7}{7}\selectfont \(\pm4.9\)}} & \mbox{37.3\textsuperscript{\fontsize{7}{7}\selectfont \(\pm5.6\)}} & \mbox{98.4\textsuperscript{\fontsize{7}{7}\selectfont \(\pm0.8\)}} & \mbox{93.6\textsuperscript{\fontsize{7}{7}\selectfont \(\pm0.7\)}} & \mbox{87.5\textsuperscript{\fontsize{7}{7}\selectfont \(\pm2.8\)}} & \mbox{89.3\textsuperscript{\fontsize{7}{7}\selectfont \(\pm0.6\)}} & \cellcolor{tablesecond}\mbox{\underline{37.2}\textsuperscript{\fontsize{7}{7}\selectfont \(\pm4.1\)}} & \mbox{91.1\textsuperscript{\fontsize{7}{7}\selectfont \(\pm3.8\)}} & 10.17 & 8.39\\
\mbox{Claude-Opus-5} & \cellcolor{tablebest}\mbox{\textbf{37.1}\textsuperscript{\fontsize{7}{7}\selectfont \(\pm1.1\)}} & \cellcolor{tablebest}\mbox{\textbf{38.8}\textsuperscript{\fontsize{7}{7}\selectfont \(\pm1.8\)}} & \cellcolor{tablebest}\mbox{\textbf{40.9}\textsuperscript{\fontsize{7}{7}\selectfont \(\pm2.2\)}} & \cellcolor{tablesecond}\mbox{\underline{99.0}\textsuperscript{\fontsize{7}{7}\selectfont \(\pm0.5\)}} & \mbox{93.8\textsuperscript{\fontsize{7}{7}\selectfont \(\pm0.9\)}} & \mbox{91.2\textsuperscript{\fontsize{7}{7}\selectfont \(\pm0.9\)}} & \mbox{89.0\textsuperscript{\fontsize{7}{7}\selectfont \(\pm1.0\)}} & \cellcolor{tablebest}\mbox{\textbf{39.8}\textsuperscript{\fontsize{7}{7}\selectfont \(\pm1.2\)}} & \mbox{91.1\textsuperscript{\fontsize{7}{7}\selectfont \(\pm10.2\)}} & 15.19 & 8.44\\
\mbox{Qwen3.8-Max} & \mbox{35.1\textsuperscript{\fontsize{7}{7}\selectfont \(\pm2.8\)}} & \mbox{37.6\textsuperscript{\fontsize{7}{7}\selectfont \(\pm2.4\)}} & \cellcolor{tablesecond}\mbox{\underline{37.9}\textsuperscript{\fontsize{7}{7}\selectfont \(\pm3.3\)}} & \cellcolor{tablebest}\mbox{\textbf{99.2}\textsuperscript{\fontsize{7}{7}\selectfont \(\pm0.0\)}} & \cellcolor{tablebest}\mbox{\textbf{97.6}\textsuperscript{\fontsize{7}{7}\selectfont \(\pm0.6\)}} & \cellcolor{tablesecond}\mbox{\underline{93.5}\textsuperscript{\fontsize{7}{7}\selectfont \(\pm3.2\)}} & \cellcolor{tablesecond}\mbox{\underline{95.4}\textsuperscript{\fontsize{7}{7}\selectfont \(\pm0.6\)}} & \cellcolor{tablesecond}\mbox{\underline{37.2}\textsuperscript{\fontsize{7}{7}\selectfont \(\pm3.3\)}} & \mbox{93.3\textsuperscript{\fontsize{7}{7}\selectfont \(\pm6.7\)}} & 5.28 & 9.26\\
\mbox{GPT-6.1-Sol} & \mbox{26.8\textsuperscript{\fontsize{7}{7}\selectfont \(\pm1.5\)}} & \mbox{30.8\textsuperscript{\fontsize{7}{7}\selectfont \(\pm1.1\)}} & \mbox{32.0\textsuperscript{\fontsize{7}{7}\selectfont \(\pm1.8\)}} & \mbox{98.4\textsuperscript{\fontsize{7}{7}\selectfont \(\pm0.8\)}} & \cellcolor{tablesecond}\mbox{\underline{97.0}\textsuperscript{\fontsize{7}{7}\selectfont \(\pm0.6\)}} & \cellcolor{tablebest}\mbox{\textbf{96.2}\textsuperscript{\fontsize{7}{7}\selectfont \(\pm0.8\)}} & \cellcolor{tablebest}\mbox{\textbf{95.9}\textsuperscript{\fontsize{7}{7}\selectfont \(\pm0.6\)}} & \mbox{29.6\textsuperscript{\fontsize{7}{7}\selectfont \(\pm1.1\)}} & \cellcolor{tablebest}\mbox{\textbf{100.0}\textsuperscript{\fontsize{7}{7}\selectfont \(\pm0.0\)}} & 6.67 & \cellcolor{tablebest}\textbf{7.76}\\
\midrule
\rowcolor{tablegroup}\multicolumn{12}{c}{\textit{Local Open-Weight Models}}\\
\midrule
\mbox{Qwen3.5-35B-A3B} & \cellcolor{tablesecond}\mbox{\underline{19.9}\textsuperscript{\fontsize{7}{7}\selectfont \(\pm0.9\)}} & \cellcolor{tablesecond}\mbox{\underline{19.8}\textsuperscript{\fontsize{7}{7}\selectfont \(\pm2.3\)}} & \cellcolor{tablesecond}\mbox{\underline{22.4}\textsuperscript{\fontsize{7}{7}\selectfont \(\pm1.8\)}} & \mbox{90.8\textsuperscript{\fontsize{7}{7}\selectfont \(\pm1.8\)}} & \mbox{81.7\textsuperscript{\fontsize{7}{7}\selectfont \(\pm1.3\)}} & \cellcolor{tablesecond}\mbox{\underline{67.1}\textsuperscript{\fontsize{7}{7}\selectfont \(\pm4.1\)}} & \mbox{68.9\textsuperscript{\fontsize{7}{7}\selectfont \(\pm0.6\)}} & \cellcolor{tablesecond}\mbox{\underline{21.1}\textsuperscript{\fontsize{7}{7}\selectfont \(\pm1.4\)}} & \mbox{60.0\textsuperscript{\fontsize{7}{7}\selectfont \(\pm11.5\)}} & -- & \cellcolor{tablesecond}\underline{8.71}\\
\mbox{Qwen3.5-122B-A10B} & \cellcolor{tablebest}\mbox{\textbf{22.7}\textsuperscript{\fontsize{7}{7}\selectfont \(\pm2.3\)}} & \cellcolor{tablebest}\mbox{\textbf{24.8}\textsuperscript{\fontsize{7}{7}\selectfont \(\pm2.3\)}} & \cellcolor{tablebest}\mbox{\textbf{27.7}\textsuperscript{\fontsize{7}{7}\selectfont \(\pm1.4\)}} & \cellcolor{tablesecond}\mbox{\underline{91.3}\textsuperscript{\fontsize{7}{7}\selectfont \(\pm0.8\)}} & \cellcolor{tablesecond}\mbox{\underline{85.3}\textsuperscript{\fontsize{7}{7}\selectfont \(\pm2.1\)}} & \mbox{65.7\textsuperscript{\fontsize{7}{7}\selectfont \(\pm3.3\)}} & \cellcolor{tablesecond}\mbox{\underline{71.3}\textsuperscript{\fontsize{7}{7}\selectfont \(\pm1.0\)}} & \cellcolor{tablebest}\mbox{\textbf{25.4}\textsuperscript{\fontsize{7}{7}\selectfont \(\pm1.9\)}} & \cellcolor{tablesecond}\mbox{\underline{71.1}\textsuperscript{\fontsize{7}{7}\selectfont \(\pm21.4\)}} & -- & 8.98\\
\mbox{Qwen3.5-397B-A17B} & \mbox{17.6\textsuperscript{\fontsize{7}{7}\selectfont \(\pm1.6\)}} & \mbox{16.7\textsuperscript{\fontsize{7}{7}\selectfont \(\pm0.8\)}} & \mbox{19.9\textsuperscript{\fontsize{7}{7}\selectfont \(\pm0.5\)}} & \cellcolor{tablebest}\mbox{\textbf{95.8}\textsuperscript{\fontsize{7}{7}\selectfont \(\pm2.0\)}} & \cellcolor{tablebest}\mbox{\textbf{87.6}\textsuperscript{\fontsize{7}{7}\selectfont \(\pm0.3\)}} & \cellcolor{tablebest}\mbox{\textbf{78.6}\textsuperscript{\fontsize{7}{7}\selectfont \(\pm0.7\)}} & \cellcolor{tablebest}\mbox{\textbf{80.3}\textsuperscript{\fontsize{7}{7}\selectfont \(\pm0.6\)}} & \mbox{18.9\textsuperscript{\fontsize{7}{7}\selectfont \(\pm1.1\)}} & \cellcolor{tablebest}\mbox{\textbf{84.4}\textsuperscript{\fontsize{7}{7}\selectfont \(\pm7.7\)}} & -- & \cellcolor{tablebest}\textbf{8.22}\\
\bottomrule
\end{tabularx}
\end{table*}

\noindent\textbf{Strong Core performance coexists with limited decision help.} Table~\ref{tab:model-results} shows this contrast under \textit{Principle-Guided CART Prompting}. This \textbf{\textit{understanding-to-execution gap}} means that meeting required behavior does not ensure high-quality purchase advice. The gap persists despite explicit Ideal Response Principles. Core specifies minimum obligations; Excellence rewards comparisons, priorities, and conditional advice. We next compare specific obligations and broad CART scenarios.

\noindent\textbf{Goal alignment exceeds requirement adherence for all 10 models and uncertainty preservation for 9} in the first full-suite evaluation. Specific conditions still require separate checks. No model passes every required check in this comparison. Safety rates separately pool case-specific risk checks.

\noindent\textbf{Constraints has the lowest Excellence rate for 7 of 10 models} in the first full-suite evaluation. These cases require recommendations that reconcile interacting constraints or dependent choices. The groups overlap. We next diagnose Core failures and examine where the same modes recur.

\subsection{Failure Mode Analysis}
\label{sec:failure-analysis}
We first identify shared Core failure modes and locate the same errors across scenarios. Separately, we assess plan validity after resolved need and evidence checks pass.

\subsubsection{Shared Failure Modes}
\label{sec:shared-failure-modes}
\begin{figure}[!t]
\centering
\includegraphics[width=\columnwidth]{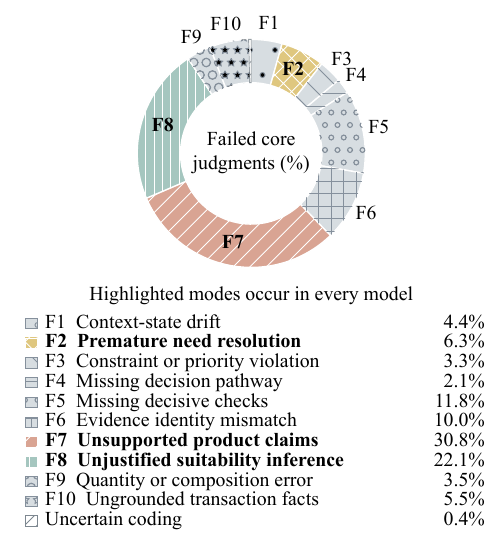}
\caption{Core failure composition. Unsupported product claims (F7), unjustified suitability inference (F8), and premature need resolution (F2) recur in every model. Slices retain uncertain diagnoses.}
\label{fig:failure-modes}
\end{figure}

\textbf{Unsupported product claims and unjustified suitability dominate the failure mix.} These errors account for most pooled Core failures (Figure~\ref{fig:failure-modes}). \textbf{\textit{F7: Product Evidence Overreach}} adds intrinsic product claims unsupported by supplied evidence. \textbf{\textit{F8: Product Fit Leap}} turns stated attributes into unjustified user-specific suitability guarantees. \textbf{\textit{F2: Shopping Uncertainty Collapse}} treats unknown preferences, possessions, purchase needs, or scenario conditions as settled. All three modes occur in every evaluated model, although cases differ. Thus, grounded shopping advice requires supported product facts and justified links to user needs.

\noindent\textbf{Diagnosis scope.} Each failed Core response--criterion instance receives one primary AI-assisted diagnosis. Excellence and Safety failures are excluded; uncertain codes remain in the denominator. However, the global mix can hide errors concentrated in particular scenarios. We next examine these profiles.

\subsubsection{Scenario-Specific Failures}
\label{sec:scenario-analysis}\label{sec:scenario-specific-failures}
\begin{figure}[!t]
\centering
\includegraphics[width=77mm]{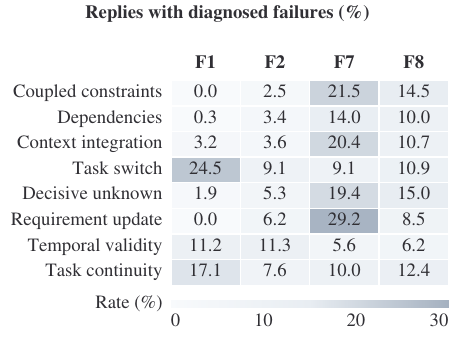}
\caption{Scenario reply rates for context-state drift (F1), premature need resolution (F2), unsupported product claims (F7), and unjustified suitability inference (F8). Cells use equal model and case weights; groups overlap.}
\label{fig:scenario-failures}
\end{figure}

\textbf{Context-state drift is minor overall but the most frequent diagnosed mode in task-switch cases.} Following the shared-mode diagnosis, Figure~\ref{fig:scenario-failures} maps F2, F7, and F8 across scenarios alongside \textbf{\textit{F1: context-state drift}}. F1 misuses a stated actor, requirement, or temporal state. Drift also appears when the shopping task continues. The continuity--F1 association remains positive after adjusting for overlapping mechanisms and rubric opportunities. Task continuity therefore does not guarantee correct state tracking. Even within one shopping goal, the relevant actor, constraints, or temporal state may change. Unlike the failure composition, scenario rates use all replies, with equal model and case weights. Groups and modes overlap.

\begin{figure}[!t]
\centering
\includegraphics[width=77mm]{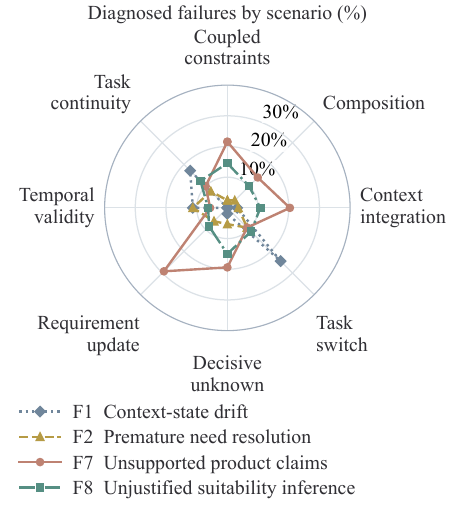}
\caption{Task switches show the most context-state drift. Radius gives failure rates among all scenario replies (0--40\%), equally weighting models and cases.}
\label{fig:scenario-radar}
\end{figure}

\textbf{Decisive unknowns expose more evidence and fit errors than premature need resolution.} In decisive-unknown scenarios, both F7 and F8 occur more often than F2. Thus, preserving unknowns also requires bounding product and suitability claims. Evidence checks ask whether product claims follow from the supplied records. Fit checks ask whether those supported attributes warrant a recommendation for the current user. Figure~\ref{fig:scenario-radar} shows the same profiles. We next test plan validity after need and evidence checks pass.

\subsubsection{Shopping Feasibility Mirage}
\label{sec:feasibility-mirage}
\textbf{Replies can meet assessed user requirements and pass product evidence checks yet fail to justify a workable recommendation.} Eligible replies need resolved evidence and plan checks, plus a resolved goal, requirement, or uncertainty check. We then select replies passing all resolved goal, requirement, uncertainty, and evidence checks. Passing an uncertainty check can mean preserving an unknown, so selected replies need not settle every missing preference or fact. Among these replies, 13.0\% fail a plan check. This estimate weights models equally (95\% interval, 8.1--18.4). The analysis concerns plan validity within Core.

\section{Conclusion}
\benchname{} pairs timestamped dialogue and product evidence with case-specific rubrics, guided by Ideal Response Principles and Principle-Guided CART Prompting. Expert-confirmed results reveal an understanding-to-execution gap and shared failures in evidence, suitability, and unresolved needs. Even replies passing all resolved need and evidence checks can fail plan checks.

\section*{Limitations}
First, although we cover common shopping scenarios such as gifting and travel clothing, specialized needs merit further study. Examples include historically accurate costumes for museum reenactments and equipment for extended polar expeditions.
Second, although we assess recommendations under changing needs, feedback from actual product use may introduce new requirements. How advice should adapt across longer purchase, return, and replacement cycles merits further study.

\section*{Ethical Considerations}
\benchname{} is provided solely for academic research and evaluation. Our cases are grounded in real commercial shopping needs. Product information is supplied by an internal product search platform. Internet searches were used only to confirm that corresponding products could be found online. Third-party content remains subject to the original rights and applicable terms. Experts annotate requirements from visible dialogue and retrieved evidence. Evaluation rewards grounded explanations and respects user choice.

\FloatBarrier

\end{document}